\documentclass[runningheads]{llncs}
\usepackage[T1]{fontenc}
\usepackage{graphicx}
\usepackage{hyperref}
\usepackage{amsmath, amssymb}
\usepackage{multirow}
\usepackage{float}
\floatstyle{ruled}
\newfloat{algorithm}{t}{loa}
\floatname{algorithm}{Algorithm}
\usepackage{color}

\begin{document}
\title{MORPH: Self-Organising Multi-Robot Task Allocation via Neuroplasticity-Inspired Adaptive Topology}
\titlerunning{MORPH: Self-Organising MRTA}
%
\author{Xuezhi Niu\orcidID{0009-0003-2381-1850} \and 
Didem G\"{u}rd\"{u}r Broo\orcidID{0000-0001-5703-5923}
}
\authorrunning{X. Niu and D. G. Broo}
%
\institute{Department of Information Technology, Uppsala University, 751 05 Uppsala, Sweden
\email{\{xuezhi.niu, didem.gurdur.broo\}@it.uu.se}\\}
\maketitle              
\begin{abstract}
 
Multi-robot task allocation (MRTA) in dynamic environments faces a fundamental tension: effective coordination requires learned structure, but that structure must adapt when conditions change. Existing methods resolve this by assuming prior task knowledge — a utility function, a cost matrix, or a trained policy making them brittle when deployed without such knowledge or when task distributions shift. We present MORPH (Multi-agent Online Rewiring through Plasticity-guided Hierarchy), a training-free MRTA framework in which global allocation quality emerges from four interacting local plasticity rules (synaptic, homeostatic, structural, and metaplasticity) applied to a directed pairwise preference matrix updated from runtime co-occurrence and task-completion feedback. MORPH requires no task model, no bid computation, and no offline training; response decisions use learned AGV-to-Picker preferences rather than a fixed proximity rule. Within the Gerkey–Mataric MRTA taxonomy, MORPH is the first method in the single-task, single-robot, instantaneous-assignment class to learn directed pairwise allocation preferences online. Evaluated on the TA-RWARE robotic warehouse benchmark ($N{=}8$--$24$ agents, four map scales, $T{=}800$ steps per episode, 5 seeds), MORPH achieves $110\%$ of all-to-all throughput at N=24 while using only $21\%$ of possible coordination links as an efficiency advantage that grows monotonically with fleet size. Under spatial task distribution shift, MORPH degrades $3\times$ less than proximity-based methods while its learned preferences remain uncorrelated with Manhattan distance. Systematic ablation confirms all four plasticity rules contribute measurably. Two allocation properties emerge without programming: cross-type preference dominance and progressive preference sparsification, mirroring the developmental refinement of biological neural circuits. Learned preferences are driven by task co-occurrence history, not spatial proximity. The implementation and experiment scripts are available at \url{https://github.com/Cyber-physical-Systems-Lab/morph\_v2}.

\keywords{multi-robot task allocation \and self-organisation \and emergent behaviour \and neuroplasticity \and online learning \and heterogeneous robots \and non-stationary environments \and preference learning}
\end{abstract}

\section{Introduction}
\label{sec:intro}
Biological neural circuits solve a problem that engineering has not; they build their own wiring diagrams from experience. A newborn cortex is not connected according to a pre-specified topology; instead, four interacting plasticity rules (Hebbian potentiation, homeostatic scaling, structural synaptogenesis and pruning, and metaplastic regulation) discover which connections are worth maintaining and which should be discarded, continuously, throughout the lifetime of the organism~\cite{hebb1949,holtmaat2009}. The result is a sparse, adaptive, task-semantic network that neither a human designer nor an optimisation algorithm could have specified in advance.

Multi-robot task allocation (MRTA) faces an analogous challenge. In robotic warehouses, search-and-rescue operations, and autonomous vehicle fleets, a heterogeneous team of agents must decide, at every moment, which robot should respond to which task request. This decision is made continuously as tasks arrive, robots complete assignments, and demand patterns shift. The quality of allocation directly determines system throughput, and failures to adapt when conditions change translate immediately into degraded performance. To this end, the core problem we address is the following: 
\begin{itemize}
  \item \textit{How can a multi-robot team learn who should coordinate with whom, primarily from runtime co-occurrence statistics, without a task model, without relying on spatial priors, and without offline training, while remaining adaptive when task distributions change?}
  \begin{itemize}
  \item RQ1 (algorithmic): What mechanisms enable the online inference and update of directed coordination preferences from co-occurrence and reward signals, in the absence of task models and spatial priors?
  \item RQ2 (adaptation and resilience): To what extent does the learned coordination structure remain stable yet responsive under non-stationary task distributions, and how does it affect system-level resilience (e.g., performance retention and recovery under shifts or perturbations)?
  \end{itemize}
\end{itemize}

A few existing MRTA methods sidestep this challenge by assuming prior task knowledge is available at deployment time. For instance, auction and market-based methods~\cite{choi2009,dias2006} require each robot to compute a score over available tasks; optimisation-based methods \cite{gerkey2004} require a cost matrix; learning-based approaches \cite{sartoretti2019,wang2021} train allocation policies end-to-end over millions of environment steps on a fixed distribution. Similarly, proximity heuristics avoid explicit task modelling but encode spatial structure as a surrogate, requiring position sensors and fixed-radius recalibration per environment. Swarm-inspired and threshold-based methods \cite{bonabeau1998,castello2016} operate without task models but produce per-robot scalar thresholds rather than directed pairwise allocation preferences, and store coordination signals in the environment rather than in agent memory. No existing method learns directed pairwise allocation preferences online from co-occurrence statistics alone, without any of these assumptions, and adapts them continuously when conditions change.

We present \textbf{MORPH} (Multi-agent Online Rewiring through Plasticity-guided Hierarchy), a training-free MRTA framework that directly imports the four biological plasticity principles into the coordination layer of a heterogeneous robot team. The contributions of this paper are as follows. We propose MORPH, the first MRTA method that learns directed pairwise allocation preferences online from runtime co-occurrence and task-completion feedback without a task model, bid computation, or offline training, and we provide a precise mapping of each plasticity rule to the allocation failure mode it prevents. We demonstrate that selective neuroplasticity-inspired coordination outperforms all-to-all communication at every tested scale. Finally, we provide mechanistic evidence that the learned preference structure is task-semantic rather than spatial, and that two biologically motivated allocation properties emerge without being programmed: a stable coordination core of highly-trusted long-term partnerships, and a dynamic exploratory periphery of continuously tested candidate links.

The remainder of this paper is organised as follows. Section~\ref{sec:related} reviews related work on MRTA, self-organising multi-agent systems, and neuroplasticity-inspired learning. Section~\ref{sec:method} describes the four MORPH plasticity rules and their biological grounding. Section~\ref{sec:experiments} presents the experimental setup and results across four complementary dimensions, such as throughput and communication efficiency, plasticity rule ablation, adaptability under task distribution shift, and emergent allocation structure. Section~\ref{sec:discussion} discusses MORPH's position in the MRTA design space and its limitations. Section~\ref{sec:conclusion} concludes.
 
\section{Related Work}
\label{sec:related}
Gleizes et al.~\cite{gleizes1999} introduced the Adaptive Multi-Agent Systems (AMAS) theory, which provides a principled foundation for systems in which global functionality emerges from local cooperative interactions without central control or explicit task models. The core principle is that agents which maintain cooperative relationships with their neighbourhood will collectively produce an adequate global function. Applications of AMAS have spanned industrial scheduling, satellite mission planning, and mobility-as-a-service coordination, as shown by Perles et al.~\cite{perles2023}, demonstrating that local cooperation rules can substitute for global optimisation in dynamic, non-stationary environments. MORPH shares this self-organising philosophy: global allocation quality emerges from local plasticity rules applied to pairwise agent relationships, with no central controller and no prescribed topology. The key distinction is that MORPH's preference matrix is a directed, continuously updated structure that accumulates task co-occurrence history (a quantitative coordination memory that reactive self-organising systems do not maintain) and that the four plasticity rules governing its evolution are grounded directly in biological mechanisms rather than derived from cooperative equilibrium conditions.
 
Market-based and auction approaches are the dominant paradigm for MRTA in heterogeneous robot teams \cite{dias2006}. Choi et al.~\cite{choi2009} introduced the Consensus-Based Bundle Algorithm, which provides convergence guarantees and handles heterogeneous agents through role-specific scoring functions, and extensions address coupled task constraints \cite{johnson2011} and asynchronous communication \cite{ponda2012}. These methods share a common structural requirement: a utility function must be specified before deployment, encoding prior knowledge about which robot-task combinations are valuable. When task co-occurrence patterns must be discovered from experience, as in environments where heterogeneous role pairings emerge at runtime, the scoring function does not exist and auction methods cannot operate. MORPH requires no such prior; its allocation preferences emerge entirely from runtime statistics. Similarly, linear assignment, MILP, and combinatorial optimisation methods produce globally optimal or near-optimal allocations given a cost matrix \cite{gerkey2004,korsah2013}. Many realistic MRTA formulations with combinatorial or coupling constraints are NP-hard and require exponential-time solvers in the worst case \cite{korsah2013}. They are therefore unsuitable for online MRTA at scale: computation grows super-linearly with fleet size, task arrivals require continuous re-solving, and the cost matrix must be known in advance. MORPH operates with $O(N^2)$ preference lookups per step and no offline computation, making it viable from the first episode in any environment.
 
Learning-based methods train allocation and coordination policies end-to-end with reinforcement learning. For instance, Sartoretti et al.~\cite{sartoretti2019} (PRIMAL) and Wang et al.~\cite{wang2021} (RODE) train allocation and role-decomposition policies over millions of environment steps, achieving strong performance on their training distribution. Dynamic coordination-graph methods make the interaction structure explicit; CommFormer learns a sparse communication graph jointly with the task policy via bi-level optimisation~\cite{hu2024}. These methods show that sparse, dynamic agent-interaction structure can improve coordination, but they still depend on trained graph encoders, value functions, or communication policies. MORPH is not competing with these methods in raw post-training throughput; its contribution is different in kind. MORPH trains no graph encoder or value function: the coordination graph itself is a plastic structure updated online from co-occurrence and reward, operating from step zero on any existing controller and adapting when the task distribution shifts.

Prior work demonstrated that bio-inspired coordination principles can improve system performance and resource utilisation in warehouse multi-robot systems using a trained CTDE framework~\cite{niu2025icps,niu2025icras}; MORPH extends this research direction by removing the training requirement entirely, enabling coordination structure to emerge online from plasticity rules rather than offline from trained optimisation objectives. In robotics more broadly, neural process models such as ROBOVERINE use biologically inspired neural dynamics to combine attention, scene grammar, and online search in naturalistic environments~\cite{grieben2024roboverine}. MORPH follows the same general design stance of using neural mechanisms for adaptive robotic behaviour, but applies them to coordination topology rather than visual search. Connecting spatially close agents via a fixed radius $r$ is the most common deployment heuristic in robot coordination~\cite{gerkey2004}. On the other hand, proximity is simple and effective when spatial closeness correlates with task co-assignment, but it requires position sensors, must be recalibrated per environment, and encodes a fixed spatial prior that cannot adapt when demand patterns change.
 
Dorigo and St\"{u}tzle~\cite{dorigo2004} developed Ant Colony Optimisation, and Salman et al.~\cite{salman2024} describe stigmergy-based approaches that coordinate through environment-mediated signals: the environment acts as the coordination medium, and coordination memory lives in pheromone traces rather than in agent-pair
weights. Response threshold models~\cite{bonabeau1998,castello2016} assign per-robot scalar thresholds that produce probabilistic task specialisation without explicit assignment. Both families operate without task models, but MORPH is fundamentally different because a directed, pairwise, continuously updated preference structure that lives in agent memory and accumulates coordination history across the episode. This distinction matters at task distribution shift: stigmergy decays when the environment changes, threshold models adjust scalar specialisations, but neither maintains a directed coordination history that can be selectively pruned and rewired by metaplastic mechanisms.

Finally, prior work on neural plasticity motivates MORPH's four-rule decomposition. Reward-modulated Hebbian learning can explain selective network reorganisation from global feedback in brain-control tasks~\cite{legenstein2010rewardmodulated} and can specialise generic recurrent microcircuits without a supervised teacher~\cite{hoerzer2014emergence}. Structural plasticity has also been modelled as binary activation and deactivation of network units, linking growth and pruning to cost-aware sparsification~\cite{li2021neural}. Dynamic threshold mechanisms have recently been used in spiking neural networks to stabilise activity and improve adaptive computation~\cite{ding2022biologically}. More directly, Zhou et al.~\cite{zhou2026dynamic} propose a BCM-inspired Dynamic Weight Adaptation Mechanism for spiking neural networks, dynamically adjusting weights to provide homeostasis under degraded continuous-control and obstacle-avoidance conditions. These works apply biologically inspired dynamic thresholds and BCM homeostasis to prevent weight dysregulation within single neural networks. MORPH translates the same principle to a qualitatively different problem, rather than stabilising activations within a network, BCM here regulates pairwise coordination preferences between agents in a multi-robot system, preventing topology inertia when task relationships dissolve and driving selective rewiring under distributional shift. Metaplastic and sampling views interpret plasticity as controlled exploration over network configurations, with temperature-like regulation connecting reinforcement learning (RL) and annealing~\cite{kappel2015Nian11Yue6Rinetwork,yu2016hamiltonian}, while recent deep-RL analysis shows why preserving plasticity matters for continued adaptation~\cite{lyle2023understanding}. Differentiable plasticity~\cite{miconi2018} and Backpropamine~\cite{miconi2019} optimise plasticity rules \emph{within} neural networks. MORPH transfers these ideas to a different object: the directed allocation preference graph \emph{between} agents, where plasticity controls coordination topology rather than internal neural representation.

\section{The MORPH Framework}
\label{sec:method}
MORPH maintains a directed preference matrix $\mathbf{W} \in [0,1]^{N \times N}$, initialised to zero at the start of each episode and updated online by four interacting plasticity rules. The entry $W_{ij}$ encodes how strongly robot~$i$ should preferentially request a response from robot~$j$, accumulated entirely from runtime co-assignment statistics. The induced allocation policy is:
\begin{equation}
  P(\text{respond}_{ij}) = W_{\text{floor}} + (1 - W_{\text{floor}}) \cdot W_{ij},
  \quad W_{\text{floor}} = 0.25
  \label{eq:policy}
\end{equation}
The floor of $0.25$ provides residual coordination capacity before preferences are established, analogous to baseline synaptic transmission in biological circuits~\cite{dayan2001}. MORPH operates as a coordination preference layer over any existing robot controller, requiring no modification to the underlying task policy.
 
Table~\ref{tab:rules} summarises the four rules, their biological basis, and the specific allocation failure each one prevents. The remainder of this section describes each rule and the implementation constraints required for biological fidelity.
 
\begin{table}[ht]
\centering
\caption{MORPH plasticity rules: biological basis and allocation failure prevented.}
\label{tab:rules}
\begin{tabular}{llll}
\hline
Rule & Plasticity form & Biological basis & Failure prevented \\
\hline
R1 & Synaptic & Hebbian LTP + dopamine gating~\cite{hebb1949,dayan2001} & Stale preferences persist \\
R2 & Homeostatic & Synaptic scaling~\cite{turrigiano2004} & Isolation / hub collapse \\
R3 & Structural & Synaptogenesis + pruning~\cite{holtmaat2009} & Cold start; slow rewiring \\
R4 & Metaplasticity & BCM-inspired dynamic thresholds~\cite{bienenstock1982,ding2022biologically,zhou2026dynamic} & Saturation; shift inertia \\
\hline
\end{tabular}
\end{table}
 
\subsection{R1: Synaptic Plasticity (every step)}
 
Hebb's rule states that synapses between co-active neurons strengthen over time~\cite{hebb1949}. Biologically, this potentiation is gated by dopamine: only co-activations that precede reward are consolidated, preventing reinforcement of coincidental activity~\cite{dayan2001}. Without synaptic plasticity, MORPH would treat every step independently, with no mechanism to distinguish reliably useful partners from incidental ones.
 
MORPH instantiates R1 as a reward-gated Hebbian update over the Jaccard task-coverage overlap $J^t_{ij}$:
\begin{equation}
  W^{t+1}_{ij} = W^t_{ij} \cdot \gamma
    + \alpha \cdot J^t_{ij} \cdot r^t \cdot A^t_{ij},
  \qquad
  J^t_{ij} = \frac{|T^t_i \cap T^t_j|}{|T^t_i \cup T^t_j|}
  \label{eq:hebbian}
\end{equation}
where $T_i^t$ and $T_j^t$ are the sets of tasks assigned to robots $i$ and $j$ at timestep $t$, $\gamma$ is the base decay rate, $\alpha$ the learning rate, $r^t$ the step reward, and $A^t_{ij} \in \{0,1\}$ the structural link indicator. The Jaccard signal ($J^t_{ij}$) encodes genuine task cooperation rather than physical proximity, the reward ($r^t$) gates consolidation so that only preferences contributing to completed deliveries are strengthened, and $A^t_{ij}$ restricts updates to active links, preventing broad and non-selective reinforcement of all co-active pairs. Only preferences with $J^t_{ij} > 0.25$ receive delivery-burst credit, mirroring the biological requirement that dopamine-gated LTP requires substantial co-activation, not mere coincident presence~\cite{dayan2001}.
 
\subsection{R2: Homeostatic Plasticity (every $k$ steps)}
 
Biological neurons regulate their total synaptic input to maintain a stable firing rate through homeostatic scaling~\cite{turrigiano2004}: when a neuron receives too little input it globally scales up its synaptic strengths; when it receives too much it scales down. Without homeostatic plasticity, MORPH agents could become isolated (receiving no task co-assignments and therefore no preference updates) or over-connected (accumulating preferences faster than can be meaningfully used).
 
Each agent $i$ maintains a homeostatic correction $H^t_i$ that biases link formation toward a target degree fraction $\rho_{\text{target}}$:
\begin{equation}
  H^{t+1}_i = H^t_i + \beta \cdot \bigl(\rho_{\text{target}} - \rho^t_i\bigr)
  \label{eq:homeostatic}
\end{equation}
where $\rho_{\mathrm{target}}\in[0,1]$ is the target degree fraction, $\rho^t_i$ is the current degree fraction of agent $i$, and $\beta$ is the adaptation rate. $H_i^t$ adjusts the effective formation threshold locally, preventing isolation and hub collapse without global coordination.
 
\subsection{R3: Structural Plasticity (every $k$ steps)}
 
Biological circuits physically grow new synaptic connections toward active neighbours and prune chronically weak ones~\cite{holtmaat2009}. Anticipatory synaptogenesis allows synapses to form before full co-activation, guided by predictive signals about likely future partners. Without structural plasticity, MORPH would be limited to adjusting preferences on a fixed link set and could not discover new coordination relationships.
 
New links form when mutual information between task-assignment histories,
amplified by a pairwise homeostatic gain $H^t_{\text{gain},ij} = h^t_i + h^t_j$
and adjusted by the neuromodulatory threshold $\theta^t_{\text{form,adj}}$,
exceeds the formation criterion. The homeostatic gain biases link formation
toward under-connected agents proportionally to the strength of the MI
evidence, and a predictive hint $\delta_{\text{pred},ij}$ is added to the
MI score to close the cold-start gap before co-assignment has occurred:
\begin{align}
A_{ij} &\leftarrow 1 \quad \text{if }
  \bigl(\mathrm{MI}(h_i^t, h_j^t) + \delta_{\text{pred},ij}\bigr)
  \cdot (1 + H^t_{\text{gain},ij})
  \geq \theta^t_{\text{form,adj}}
  \text{ and } A_{ij}=0,\; i\neq j \\
A_{ij} &\leftarrow 0 \quad \text{if }
  W_{ij} < \theta_{\text{prune}}
  \text{ and } \mathrm{age}_{ij} \geq g
\end{align}
The grace period $g$ mirrors the biological consolidation window: nascent synapses are protected during establishment before becoming eligible for depression~\cite{holtmaat2009}. Applying BCM to newly formed links sets $\theta^{\text{bcm}}_{ij} = 0$, causing $W_{ij}/\theta^{\text{bcm}}_{ij} \to \infty$ and immediate maximum extra decay, killing nascent preferences before they can
establish, in direct analogy to over-depression of silent
synapses~\cite{holtmaat2009}. BCM is therefore applied only after $\mathrm{age}_{ij} > g = 20$ steps.
 
\subsection{R4: Metaplasticity (every step and every $k$ steps)}
 
Metaplasticity, the plasticity of plasticity itself, regulates how easily synaptic changes can be induced, keeping preferences within a functional dynamic range and enabling the network to shift between exploratory and consolidating modes depending on behavioural context~\cite{bienenstock1982}. R4 operates at two spatial scales.
 
\textbf{The Bienenstock--Cooper--Munro inspired sliding threshold:} maintains a per-synapse sliding modification threshold~\cite{bienenstock1982}.
When a preference consistently exceeds its own historical average, the threshold rises, making further potentiation harder. Without BCM, high-reward AGV--Picker preferences would saturate and resist pruning even after the task relationships they encoded had dissolved, creating topology inertia precisely when adaptation is most needed. Each link maintains $\theta^{\text{bcm}}_{ij}$ tracking its weight history via exponential moving average ($\tau = 0.95$). This implements the functional intent of the BCM rule as a discrete exponential moving average rather than the full differential-equation dynamics; the biological principle is preserved and adapted to the discrete multi-agent setting. Links where $W_{ij} \gg \theta^{\text{bcm}}_{ij}$ receive additional decay, applied only to mature links ($\mathrm{age}_{ij} > g$). $\theta^{\text{bcm}}_{ij}$ must be updated \emph{before} R1 fires, not after. Updating after caused neuromodulation-boosted weights to appear over-potentiated to BCM, producing extra decay that cancelled the exploration benefit from R4b, causing the two metaplasticity components to work against each other. The correct ordering ensures BCM reflects the weight state prior to the current Hebbian update, as in the biological BCM formulation~\cite{bienenstock1982}. 
 
\textbf{Neuromodulatory arousal (network level):} Neuromodulators such as dopamine
and acetylcholine shift entire circuits between exploratory and consolidating modes
depending on whether current performance meets expectation~\cite{dayan2001}. A global
arousal signal $\eta^t$ compares delivery rate against a running baseline:
\begin{equation}
  \eta^t = \frac{\hat{d}^t - d_{\exp}}{d_{\exp}}, \quad \eta^t \in [-1, +1]
  \label{eq:neuromod}
\end{equation}
When $\eta > 0$ (above expectation), effective $\alpha$ scales up, consolidating the
current preference structure. When $\eta < 0$ (below expectation), $\theta_{\text{form}}$
lowers, promoting new link exploration. This is the primary mechanism for accelerating
preference recovery after task distribution shifts.

\begin{algorithm}[htbp]
\caption{MORPH Update Cycle}
\label{alg:morph}
\small
\begin{enumerate}
\item Initialise $\mathbf{A}^0=\mathbf{0}$, $\mathbf{W}^0=\mathbf{0}$,
      $\mathbf{h}^0=\mathbf{0}$, and $\mathrm{age}=\mathbf{0}$.
\item For each timestep $t$:
  compute $J^t_{ij}$ from task co-assignments and receive $r^t$.
\item \textbf{R4a:} update the BCM threshold before Hebbian learning,
      $\theta^{\text{bcm}}_{ij}\leftarrow
      \tau\theta^{\text{bcm}}_{ij}+(1-\tau)W_{ij}$.
\item \textbf{R1:} update directed preferences,
      $W_{ij}\leftarrow W_{ij}\gamma+\alpha J^t_{ij}r^tA_{ij}$.
\item \textbf{R4b:} update $\eta^t$ and modulate $\alpha$ and
      $\theta_{\text{form}}$.
\item If $t\bmod k=0$, form links satisfying
      $(\mathrm{MI}(h_i^t,h_j^t)+\delta_{\text{pred},ij})
      (1+H^t_{\text{gain},ij})\geq\theta^t_{\text{form,adj}}$,
      prune links with $W_{ij}<\theta_{\text{prune}}$ and
      $\mathrm{age}_{ij}\geq g$, and update
      $h_i\leftarrow h_i+\beta(\rho_{\text{target}}-\rho_i)$.
\item Increment $\mathrm{age}_{ij}$ for all active $(i,j)\in E^t$ and
      continue to the next timestep.
\end{enumerate}
\end{algorithm}
 
The four rules operate on two timescales. R1 (synaptic) and R4 (neuromodulation and BCM) fire every step; R2 (homeostatic) and R3 (structural) fire every $k = 3$ steps. This separation is not merely biologically motivated. It is functionally necessary. If synaptic weight updates and structural topology changes occurred simultaneously, the preference structure would destabilise. This means that a newly formed link immediately receiving a large Hebbian update could trigger premature BCM decay before the link has established any coordination history. The two-timescale design mirrors the biological separation of fast Hebbian weight changes (millisecond to minute timescales) from axon growth and pruning (hours to days)~\cite{holtmaat2009}, and serves the same functional purpose in both settings. Algorithm~\ref{alg:morph} gives the complete update cycle; specifically the ordering of lines~4 and~5 implements the BCM-before-Hebbian constraint.

\section{Experiments and Results}
\label{sec:experiments}
\subsection{Experimental Setup}
\label{sec:setup}

\textbf{Environment.} All experiments use TA-RWARE~\cite{krnjaic2024}, a robotic warehouse benchmark in which Automated Guided Vehicles (AGVs) transport shelves to delivery stations and Picker agents assist at pickup locations. MORPH operates as a coordination preference layer over the Closest-Target Assignment (CTA) heuristic (nearest-shelf assignment with A* pathfinding), requiring no modification to the underlying controller.


\begin{figure}[ht]
  \centering
  \includegraphics[width=\textwidth]{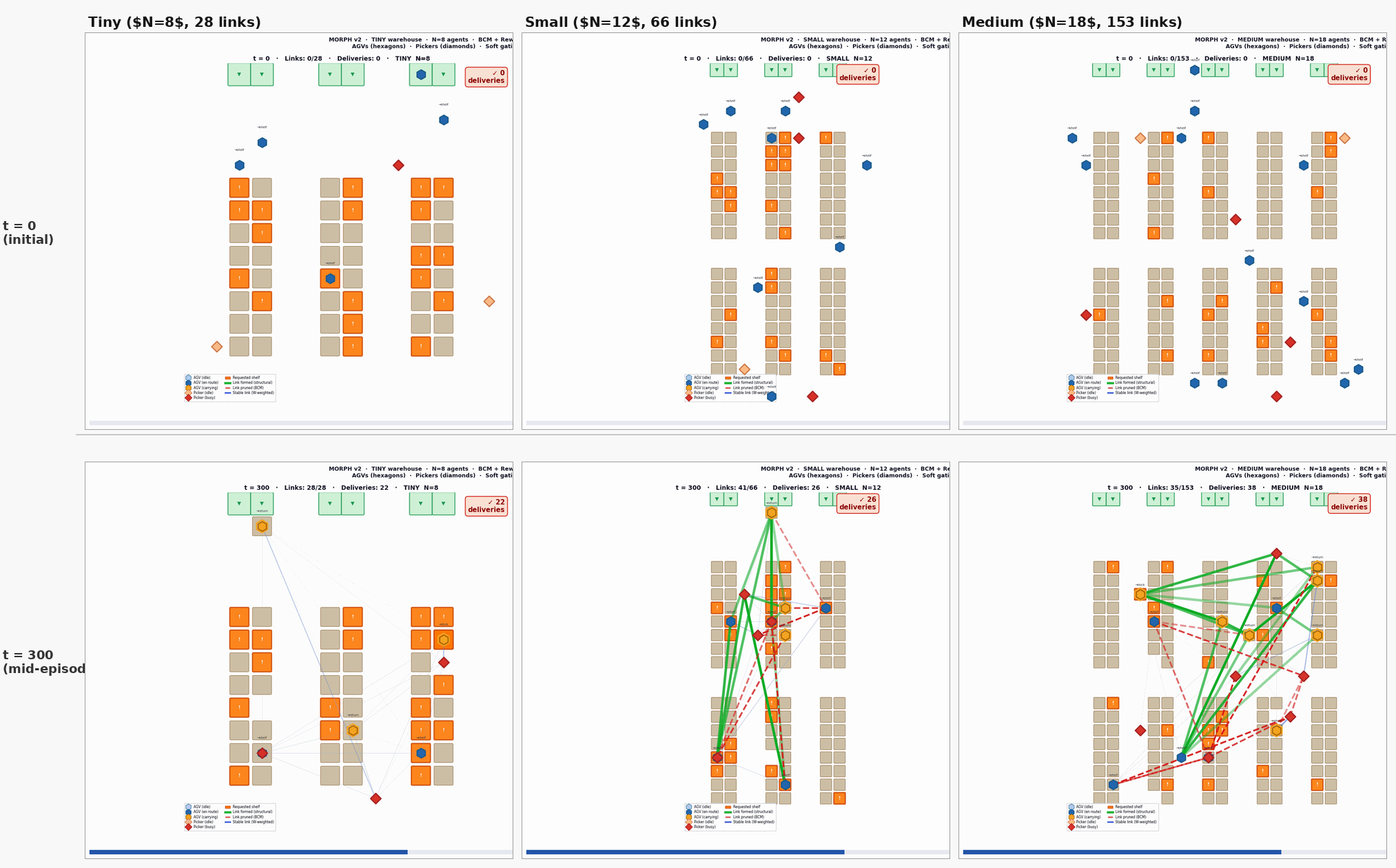}
  \caption{TA-RWARE environment simulation at three fleet scales. Top row: initial state ($t=0$) showing warehouse layout, shelf columns (orange squares), AGVs (hexagons), and Pickers (diamonds) before any preferences are established. Bottom row: mid-episode state ($t=300$) showing MORPH's self-organised coordination topology --- green lines are stable W-weighted links between AGV--Picker pairs, red dashed lines are links being pruned by thresholds. Coordination complexity grows with fleet size. Full animations are available as supplementary material.}
  \label{fig:environment}
\end{figure}

The scale study targets the core coordination bottleneck: candidate pairwise coordination links grow quadratically with fleet size, so all-to-all communication becomes increasingly noisy and expensive. We evaluate four fleet scales: Tiny ($N=8$: 5 AGVs + 3 Pickers, 28 possible links), Small ($N=12$: 8+4, 66 links), Medium ($N=18$: 12+6, 153 links), and Large ($N=24$: 16+8, 276 links), as illustrated in Figure~\ref{fig:environment}. These settings test whether MORPH becomes more selective as coordination complexity increases. The top row of Figure~\ref{fig:environment} shows the warehouse layout at initialisation; the bottom row shows the same warehouses simulation at $t=300$, by which point MORPH has established a sparse set of stable AGV--Picker links (green) while continuously exploring and pruning candidate links (red dashed). Full animations for all four scales are provided as supplementary material.
 
\textbf{Baselines:} Each baseline represents a different answer to the coordination-structure problem. \emph{Full-Graph} connects all agents to all agents ($A_{ij}=1$ for all $i,j$), testing whether dense communication is beneficial or whether it introduces coordination noise despite maximum connectivity. \textit{Proximity-$r$} connects agent pairs whose Manhattan distance is at most $r$, testing the strongest fixed spatial prior; $r$ is calibrated per scale to match MORPH's converged mean link count, intentionally giving Proximity the maximum possible link-budget advantage despite this calibration being unavailable at real deployment time. \textit{TSG (Task-Similarity Gate)} connects pairs sharing a current target location, using the same co-assignment signal as MORPH's Jaccard term but with zero memory or plasticity, isolating whether instantaneous task similarity is enough without learned preference history. 

\begin{table}[ht]
\centering
\caption{MORPH parameter values, identical across all four scales.}
\label{tab:params}
\begin{tabular}{lll}
\hline
Parameter & Symbol & Value \\
\hline
Synaptic learning rate      & $\alpha$                     & 0.18  \\
Base decay rate             & $\gamma$                     & 0.98  \\
Homeostatic rate            & $\beta$                      & 0.04  \\
Formation threshold (start/end) & $\theta_{\text{form}}^{0/T}$ & 0.75 / 0.45 \\
Pruning threshold           & $\theta_{\text{prune}}$      & 0.008 \\
Target degree fraction      & $\rho_{\text{target}}$       & 0.35  \\
Grace period                & $g$                          & 20 steps \\
Structural update frequency & $k$                          & 3 steps \\
Sliding-threshold EMA smoothing           & $\tau$                       & 0.95  \\
Reward learning rate        & $\alpha_r$                   & 0.08  \\
Neuromod gain / explore     & ---                          & 0.4 / 0.10 \\
\hline
\end{tabular}
\end{table}

\textbf{Ablation conditions:} The ablations test whether MORPH's behaviour comes from the interaction of all four rules rather than from a single heuristic component. \textit{MORPH v1} includes only R1, R2, and R3 with all v2 metaplasticity disabled, testing the value of metaplastic regulation as a whole. \textit{v2$-$threshold} removes the sliding threshold from R4, leaving neuromodulation active, testing whether saturation control is needed. \textit{v2$-$Reward} removes the delivery reward gate from R1, leaving pure Hebbian updates, testing whether co-occurrence must be tied to successful deliveries. \textit{v2$-$Neuromod} removes the neuromodulatory arousal signal from R4, leaving threshold active, testing whether global exploration--consolidation is required. \textit{MORPH v2 (full)} is the complete system. All parameter values are held fixed across all four scales and all experimental conditions; Table~\ref{tab:params} reports the full set. Statistical comparisons use Welch's two-sided $t$-test across 5 seeds; we report exact $p$-values and flag significance at $0.05$ level.
 
\subsection{Results}
\label{sec:results}

\begin{table}[ht]
\centering
\caption{TA-RWARE scale study ($T=800$, 5 seeds, mean~$\pm$~std).
\%AP = deliveries as \% of Full Graph throughput. Links = mean active
links (\% of maximum possible). Proximity $r$ calibrated per scale to
match MORPH's converged link count, giving Proximity the maximum
possible link-budget advantage.}
\label{tab:main}
\begin{tabular}{llccc}
\hline
Scale ($N$) & Method & Deliveries & \%AP & Links (\%max) \\
\hline
\multirow{4}{*}{Tiny (8)}
 & Proximity  & $62.4 \pm 4.0$  & 108\% & 22.2 (79\%)  \\
 & TSG        & $55.2 \pm 5.8$  & 96\%  & 2.6 (9\%)    \\
 & \textbf{MORPH} & $\mathbf{62.4 \pm 2.9}$ & \textbf{108\%} & 27.0 (97\%) \\
 & Full Graph  & $57.8 \pm 18.0$ & 100\% & 28.0 (100\%) \\
\hline
\multirow{4}{*}{Small (12)}
 & Proximity  & $75.2 \pm 1.6$  & 117\% & 36.2 (55\%)  \\
 & TSG        & $70.9 \pm 2.1$  & 111\% & 3.7 (6\%)    \\
 & \textbf{MORPH} & $\mathbf{70.4 \pm 1.3}$ & \textbf{110\%} & 38.0 (58\%) \\
 & Full Graph  & $64.1 \pm 13.7$ & 100\% & 66.0 (100\%) \\
\hline
\multirow{4}{*}{Medium (18)}
 & Proximity  & $102.2 \pm 2.2$ & 115\% & 30.7 (20\%)  \\
 & TSG        & $89.9 \pm 1.8$  & 101\% & 5.6 (4\%)    \\
 & \textbf{MORPH} & $\mathbf{91.0 \pm 5.1}$ & \textbf{102\%} & 40.5 (26\%) \\
 & Full Graph  & $88.8 \pm 5.9$  & 100\% & 153.0 (100\%) \\
\hline
\multirow{4}{*}{Large (24)}
 & Proximity  & $113.2 \pm 3.1$ & 116\% & 59.8 (22\%)  \\
 & TSG        & $97.6 \pm 13.5$ & 100\% & 7.8 (3\%)    \\
 & \textbf{MORPH} & $\mathbf{107.4 \pm 3.4}$ & \textbf{110\%} & 57.0 (21\%) \\
 & Full Graph  & $98.0 \pm 7.7$  & 100\% & 276.0 (100\%) \\
\hline
\end{tabular}
\end{table}

We report the results in the same order as the experimental questions: scale and communication efficiency, plasticity-rule ablation, adaptation under task distribution shift, and the emergent structure of the learned preference matrix.

\begin{figure}[!ht]
  \centering
  \includegraphics[width=\textwidth]{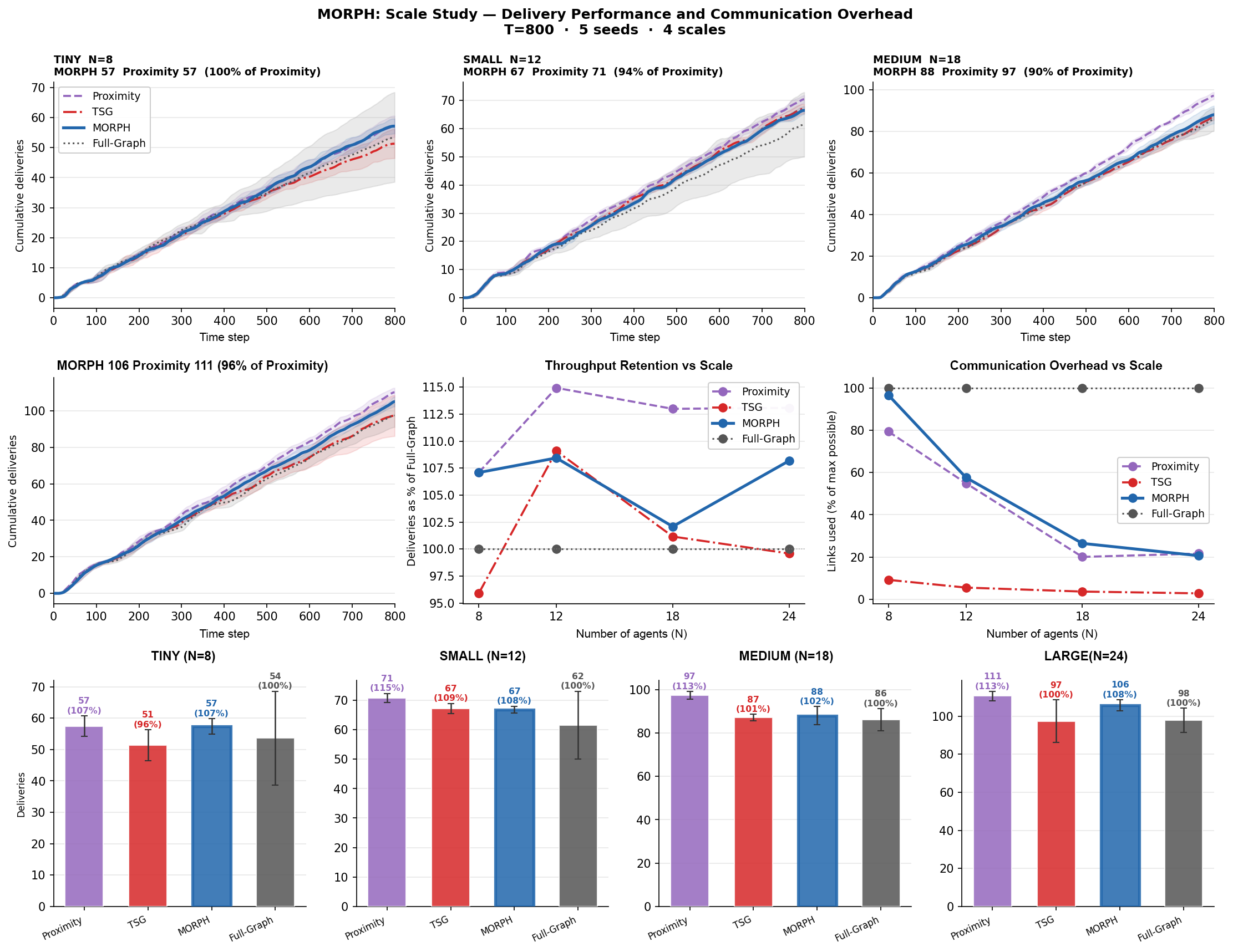}
  \caption{TA-RWARE scale study ($T=800$, 5 seeds, 4 scales). Top row: cumulative delivery curves (mean $\pm$ std shaded). Middle row: throughput retention as \% of Full-Graph (left) and links used as \% of maximum possible (right). Bottom row: grouped final-delivery bars at each scale. MORPH (blue) exceeds Full-Graph (grey) at every scale; link usage falls monotonically from 97\% to 21\% of maximum as $N$ grows from 8 to 24.}
  \label{fig:scale_comparison}
\end{figure}

\paragraph{Throughput and communication efficiency:} Table~\ref{tab:main} and Figure~\ref{fig:scale_comparison} present the main scale study results. MORPH outperforms Full Graph at every scale: At $N=24$, MORPH reaches $110\%$ of Full Graph throughput (107.4 vs. 98.0 deliveries) while using only $21\%$ of the possible links. The same pattern is visible at $N=18$, where MORPH obtains $102\%$ of Full Graph throughput with $26\%$ of the links. At the smallest scale, MORPH remains close to a dense topology (27.0 of 28 possible links), but the active-link fraction decreases from $97\%$ at $N=8$ to $58\%$, $26\%$, and $21\%$ as $N$ increases to 12, 18, and 24. Thus, MORPH does not simply approximate all-to-all coordination; it progressively concentrates coordination on a smaller fraction of pairwise relationships as the link budget grows quadratically with team size.

Proximity leads MORPH at Medium and Large ($p=0.008$ and $p=0.038$, Welch's $t$-test). Proximity benefits from the correct spatial inductive bias from step zero. In a warehouse, spatially close agents genuinely tend to co-assign. MORPH starts with zero preferences and must discover task-relevant structure from co-assignment statistics over approximately 50--100 steps. Importantly, Proximity's $r$ was calibrated from MORPH's own converged link count, a procedure unavailable at real deployment time; despite this maximum advantage, MORPH still remains competitive, reaching 70.4 vs. 75.2 deliveries at $N=12$, 91.0 vs. 102.2 at $N=18$, and 107.4 vs. 113.2 at $N=24$. The result indicates that spatial priors can be highly effective when they are correctly specified, but MORPH approaches this performance without using spatial assumptions, task models, or offline tuning.
 
TSG, which uses the same task co-assignment signal as MORPH but strips all four plasticity rules, achieves only 2--7 active links across scales. The MORPH--TSG delivery gap (9.8 deliveries at $N=24$: 107.4 vs 97.6) directly quantifies what accumulated preference memory adds over raw task-awareness alone. TSG is equivalent to a system that dissolves all preferences between assignments; no long-term potentiation, no homeostasis, no structural memory. The stable preference structure accumulated by MORPH's four interacting rules, not task-awareness alone, produces the performance gap.

\begin{figure}[ht]
  \centering
  \includegraphics[width=\textwidth]{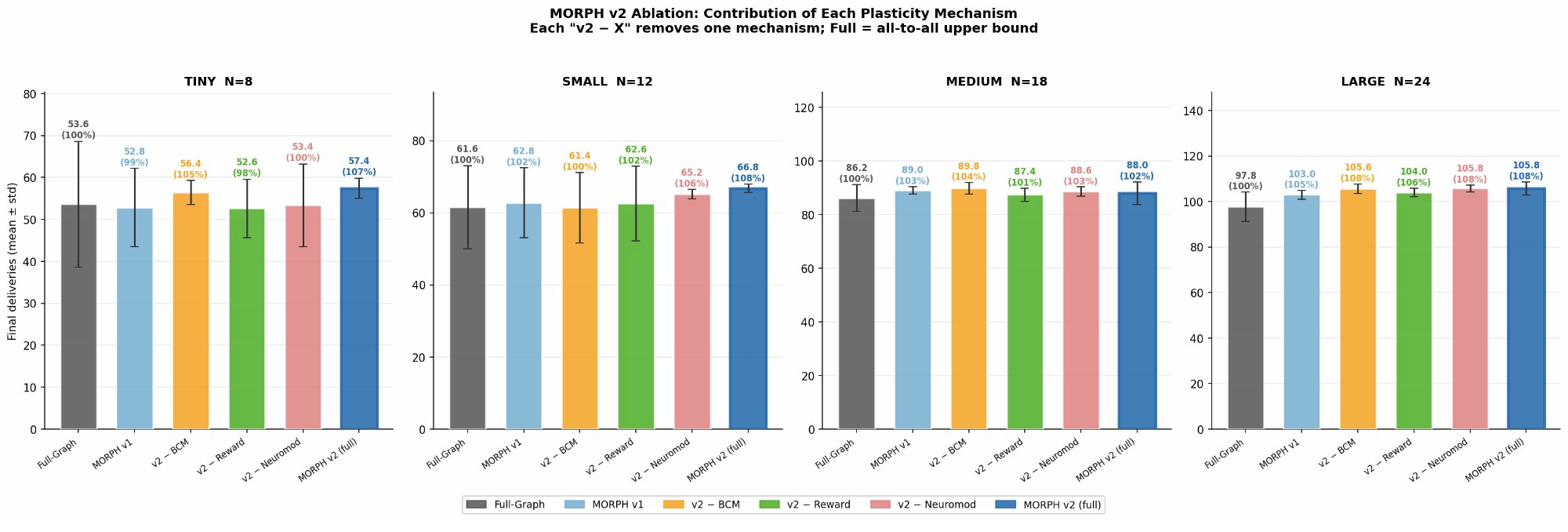}
  \caption{Plasticity rule ablation (4 scales, 5 seeds). Each ''v2$-$X'' bar removes one mechanism from the full MORPH v2 system. MORPH v2 full (dark blue) leads all partial combinations at every scale. Removing neuromodulation (pink) produces the most consistent degradation; removing the reward gate (green) causes a 7-deliveries reduction at Small.}
  \label{fig:ablation}
\end{figure}

Figure~\ref{fig:ablation} further decomposes this advantage by removing individual plasticity mechanisms from the full system. Removing Sliding-threshold (v2$-$BCM) increases performance variance without a consistent mean-throughput penalty. Without the sliding threshold, over-potentiated preferences saturate and resist pruning after task changes, producing occasional episodes where stale links dominate. Removing the delivery reward gate from R1 (v2$-$Reward, leaving pure Hebbian) causes an 8-point drop at Small (110\%$\to$102\%): strengthening all co-active preferences regardless of delivery outcome introduces noise that dilutes the signal from genuinely productive partnerships. Removing neuromodulation (v2$-$Neuromod) produces the most consistent degradation across all four scales, confirming that the global exploration--consolidation signal is the mechanism most directly responsible for performance recovery under changing conditions. MORPH v1 (R1+R2+R3, no metaplasticity) underperforms the full system at all scales, establishing that R4 adds measurable value beyond the first three rules.

\begin{figure}[ht]
  \centering
  \includegraphics[width=\textwidth]{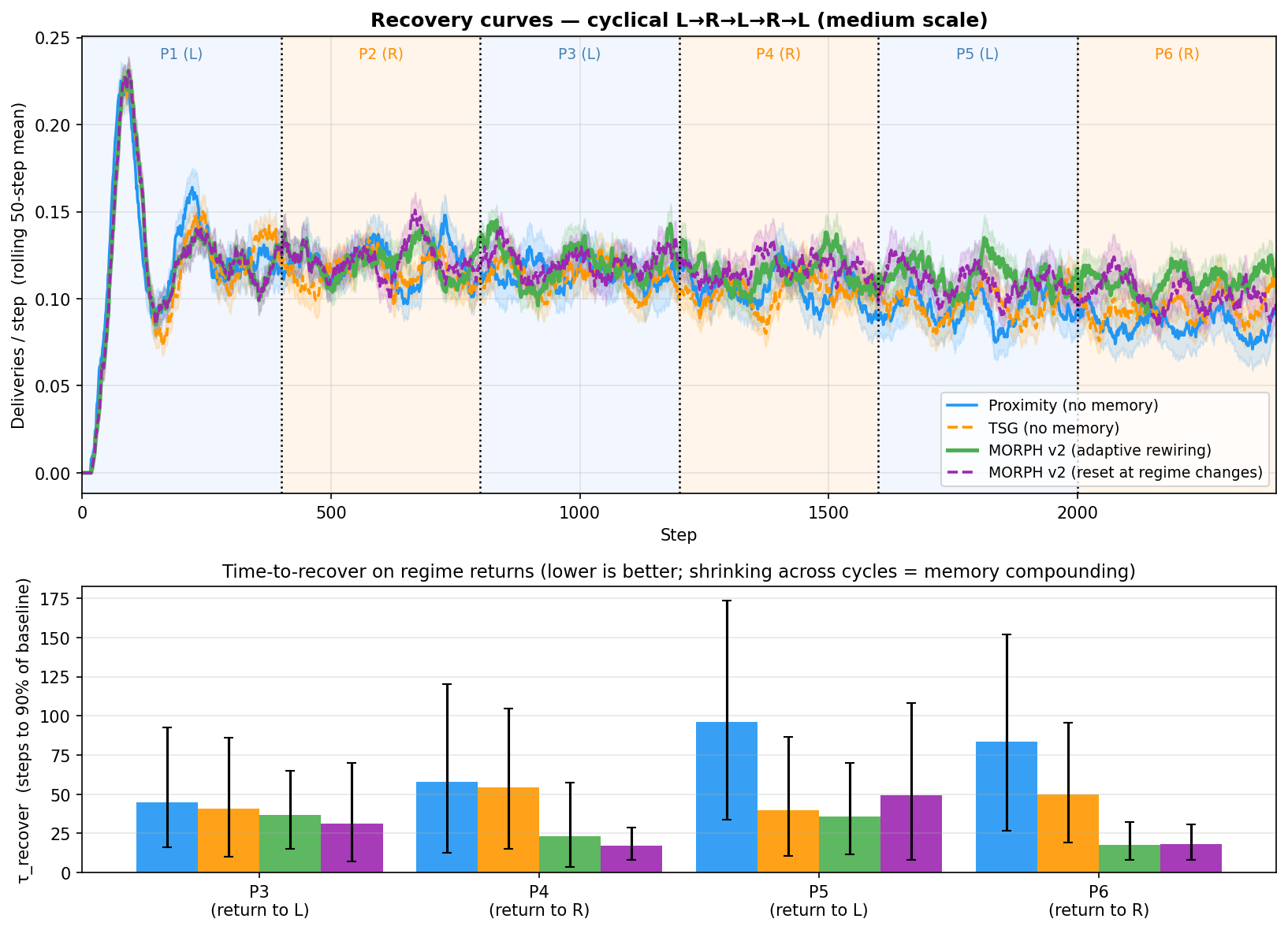}
  \caption{Adaptability under cyclical spatial task-distribution shifts at Medium scale
($N = 18$, 20 seeds). Demand alternates across six 400-step phases (L--R--L--R--L--R),
with dotted lines marking phase boundaries. \textit{Top:} Delivery rate (rolling 50-step mean)
across all six phases. \textit{Bottom:} Recovery time on repeated regime returns (steps to
reach 90\% of baseline delivery rate), where lower values indicate faster recovery and a
shrinking trend indicates memory compounding across cycles. From Phase~3 to Phase~6,
MORPH v2 shows the smallest degradation ($-7.4\%$), compared with Proximity ($-25.4\%$),
TSG ($-14.6\%$), and MORPH v2 reset ($-15.6\%$).}
  \label{fig:shift}
\end{figure}

\paragraph{Adaptability under task distribution shift:} Table~\ref{tab:shift} and Figure~\ref{fig:shift} report the task distribution shift experiment. The warehouse runs for six 400-step phases at Medium scale under an alternating left/right demand packages (L--R--L--R--L--R). Rather than measuring only total deliveries before and after one shift, we instead examine how quickly each method recovers after a regime returns. Between phases, MORPH carries its learned pairwise weights forward across all phases, so preferences strengthened during an earlier left or right demand phase remain available when the same regime reappears. MORPH-reset uses the same update rules but clears the learned weights at every phase boundary, forcing the system to relearn coordination structure from scratch after each shift. Proximity does not maintain learned preferences; it recomputes links from the agents' current spatial configuration. TSG is even more transient, rebuilding only the instantaneous co-assignment graph and discarding any accumulated evidence between assignments. For each return phase, we record the number of timesteps required for the delivery rate to recover to 90\% of the corresponding baseline performance after the shift event.

\begin{table}[htp]
\centering
\footnotesize
\caption{Task-distribution shift results at Medium scale ($N=18$, 20 seeds). P1--P6 denote mean deliveries in six consecutive 400-step phases under alternating left/right package demand. $\Delta_{6-3}$ reports the change from Phase~3 to Phase~6. r3--r6 report recovery timesteps for the corresponding return phases, where lower values indicate faster recovery.}
\label{tab:shift}
\begin{tabular}{lrrrrrrrrrrr}
\hline
Condition & P1 & P2 & P3 & P4 & P5 & P6 & $\Delta_{6-3}$ & r3 & r4 & r5 & r6 \\
\hline
Proximity & 51.1 & 48.2 & 45.7 & 41.8 & 37.9 & 34.1 & $-11.6$ & 45.0 & 57.9 & 96.0 & 83.5 \\
TSG & 49.9 & 46.0 & 44.6 & 41.4 & 38.8 & 38.1 & $-6.5$ & 40.8 & 54.6 & 39.8 & 50.0 \\
\textbf{MORPH} & 50.0 & \textbf{49.1} & 47.5 & 46.1 & \textbf{45.2} & \textbf{44.0} & $\mathbf{-3.5}$ & 36.9 & 23.1 & \textbf{35.7} & \textbf{17.7} \\
MORPH (reset at shift) & 50.0 & 48.9 & \textbf{48.2} & \textbf{46.5} & 43.9 & 40.8 & $-7.5$ & \textbf{31.4} & \textbf{17.1} & 49.5 & 18.1 \\
\hline
\end{tabular}
\end{table}

Across the six phases, the delivery counts show two effects: all methods begin from a similar level, but only MORPH maintains performance as demand regimes repeat. In Phase~1, all conditions are close, with Proximity highest at 51.1 deliveries and MORPH at 50.0. After the first shift to right-side demand, MORPH has the highest Phase~2 throughput (49.1), slightly above MORPH-reset (48.9), Proximity (48.2), and TSG (46.0). When the system returns to left-side demand in Phase~3, MORPH-reset obtains the highest mean throughput (48.2), while MORPH remains close at 47.5; this suggests that resetting can be competitive immediately when the task structure is still easy to relearn. The same pattern appears in Phase~4, where MORPH-reset is marginally highest (46.5) and MORPH follows closely (46.1). The difference becomes clearer in the later repeated regimes. In Phase~5, MORPH becomes the best-performing condition with 45.2 deliveries, while MORPH-reset drops to 43.9, TSG to 38.8, and Proximity to 37.9. In Phase~6, MORPH again leads with 44.0 deliveries, compared with 40.8 for MORPH-reset, 38.1 for TSG, and 34.1 for Proximity. Thus, the main advantage of MORPH is not only immediate recovery after a shift, but reduced cumulative degradation across repeated shifts. From Phase~3 to Phase~6, MORPH decreases by only 3.5 deliveries, compared with 11.6 for Proximity, 6.5 for TSG, and 7.5 for MORPH-reset. Relative to Phase~3, this corresponds to a 7.4\% reduction for MORPH, versus 25.4\% for Proximity, 14.6\% for TSG, and 15.6\% for MORPH-reset. Persistent MORPH weights therefore reduce long-horizon degradation by approximately $3.3\times$ compared with Proximity and $2.1\times$ compared with resetting MORPH at every phase boundary.

The recovery-time results are consistent with this interpretation, but should be read as supporting evidence because the variance is large. MORPH recovery time decreases from 36.9 steps at r3 to 17.7 steps at r6, indicating faster recovery as regimes repeat. MORPH-reset is faster in the early returns, with 31.4 and 17.1 steps at r3 and r4, but its delivery performance degrades more strongly in later phases, ending at 40.8 deliveries in Phase~6 compared with 44.0 for MORPH. This contrast suggests that resetting can produce fast short-term relearning, while persistent MORPH weights provide better retention and higher throughput under repeated regime changes. When demand shifts, obsolete AGV--Picker pairs lose Jaccard support, so R1 no longer sustains their weights. The sliding threshold accelerates decay of stale preferences, while metaplasticity lowers the formation threshold after delivery-rate drops, allowing R3 to form new zone-relevant links and homeostasis to rebalance degrees. The policy floor preserves minimal responsiveness during rewiring. In contrast, Proximity recomputes spatial links without task memory, and TSG reacts only to current co-assignment, explaining their larger long-term degradation and less stable recovery. This supports the interpretation that MORPH adapts through small, persistent connectivity updates rather than topology replacement.

\paragraph{Emergent allocation structure:} We run a long-horizon episode ($T=1500$, Medium, $N=18$, seed=42) as mechanistic evidence of the allocation structure MORPH learns at runtime. We distinguish properties attributable to MORPH's plasticity rules from those that reflect the underlying CTA heuristic and warehouse task structure.

\begin{figure}[ht]
  \centering
  \includegraphics[width=\textwidth]{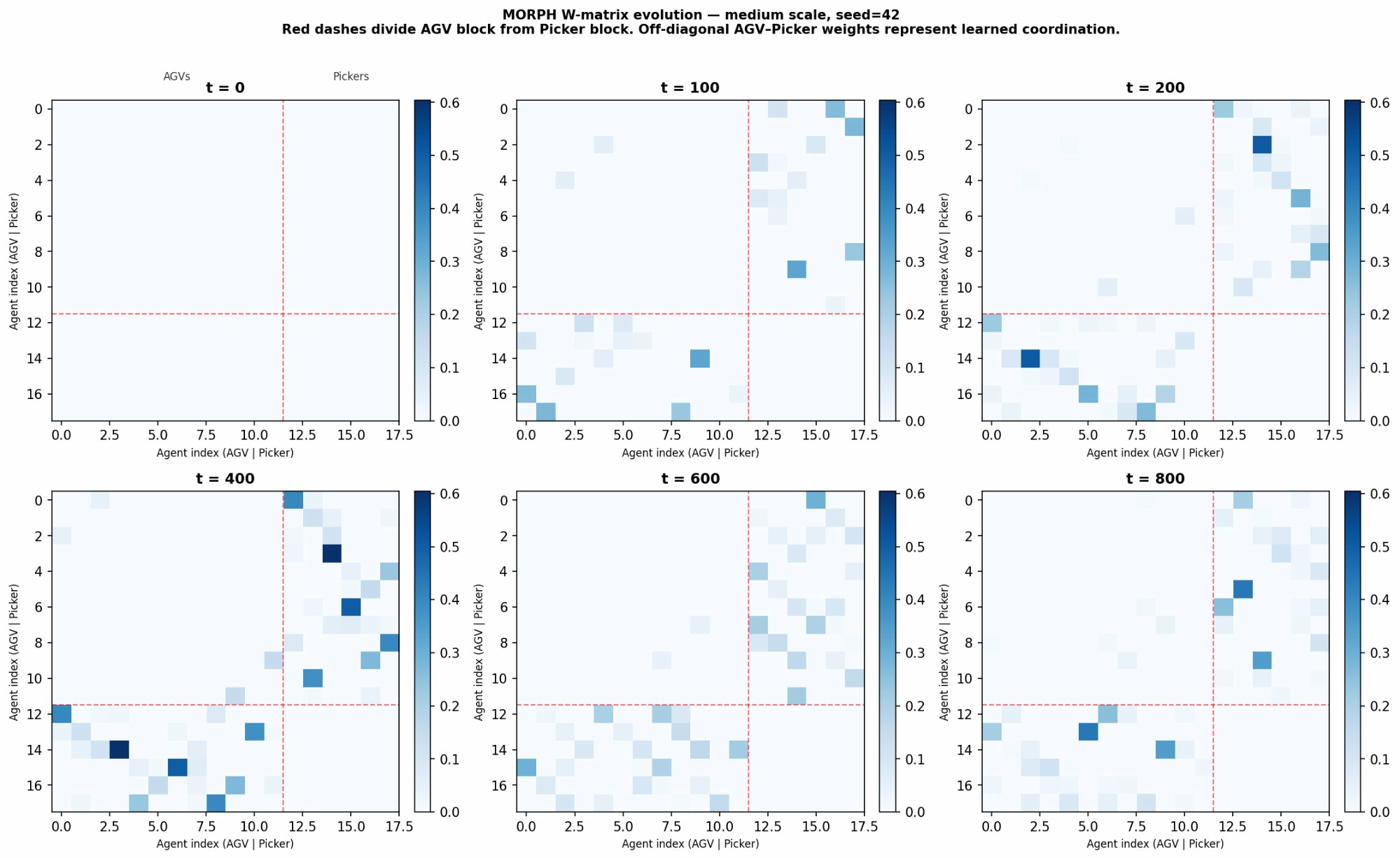}
  \caption{$\mathbf{W}$ matrix evolution ($N=18$, seed=42, snapshots at $t=0,100,200,400,600,800$). Red dashed lines divide the AGV block (rows/columns 0--11) from the Picker block (12--17). By $t=100$ the AGV--Picker off-diagonal block is already structured; diagonal blocks remain near zero throughout, consistent with the task structure.}
  \label{fig:wevolution}
\end{figure}

Figure~\ref{fig:wevolution} shows the $\mathbf{W}$ matrix evolving from all-zeros at $t=0$ to a structured, sparse pattern by $t \approx 200$. The AGV--Picker off-diagonal block develops strong weights while the AGV--AGV and Picker--Picker diagonal blocks remain near zero throughout. This is consistent with the task structure in which AGVs and Pickers cooperate on every delivery while same-type agents rarely co-assign. This cross-type preference dominance cannot arise from memoryless baselines such as TSG, which maintain no persistent preferences between assignments. The $\mathbf{W}$ matrix is therefore a learned, interpretable summary of the system's coordination history: $W_{ij}$ encodes how much agents $i$ and $j$ have historically co-assigned on deliveries. Crucially, this learned structure is not a proxy for spatial proximity. Across the $18 \times 17 = 306$ directed AGV--Picker pairs, the Spearman correlation between final $W_{ij}$ and mean Manhattan distance is $\rho = -0.111$ ($p = 0.174$, n.s.). Statistically indistinguishable from zero. Strong AGV--Picker preferences appear at all distances with no detectable decay, confirming that MORPH discovers a genuinely task-semantic coordination structure without any spatial prior, and that the cross-type block structure visible in Figure~\ref{fig:wevolution} is driven by role complementarity rather than physical arrangement.
 
The neuromodulatory signal $\eta^t$ mirrors the dopaminergic arousal dynamics described in biological reward circuits~\cite{dayan2001}. Throughout the $T=1500$ episode, $\eta$ is mildly negative during steady operation (mild exploration mode), punctuated by short consolidation bursts ($\eta > 0$) that follow clusters of successful deliveries. Active link count rises steadily through the first 200 steps as structure stabilises under this exploration--consolidation cycle, then fluctuates in a narrow band around 35--40 links as R4 continuously rebalances the preference topology. Finally, the long-horizon run reveals a stable core versus dynamic periphery structure that mirrors the brain's maintenance of long-term potentiated synapses alongside ongoing synaptic turnover~\cite{holtmaat2009}. Over 800 steps and 153 possible links, MORPH generates 772 link formation and dissolution events. Only 6 of these (0.8\%) survive $\geq 400$ consecutive steps. This is the stable coordination core of highly-trusted, repeatedly confirmed AGV--Picker partnerships. The remaining 766 events (99.2\%) are short-lived exploratory probes with a median lifetime of 21 steps. Progressive preference sparsification is further confirmed by the Fiedler value $\lambda_2$ declining monotonically from 0.57 to 0.23 over 1500 steps as sliding-threshold and structural pruning eliminate low-value preferences, mirroring the developmental refinement of maturing biological circuits~\cite{holtmaat2009}. Together, these properties (task-semantic structure, role-driven rather than proximity-driven preferences, biologically plausible neuromodulation dynamics, and a stable core surrounded by a dynamic exploratory periphery) emerge without being programmed, arising solely from the interaction of the four plasticity rules.

 
\section{Discussion}
\label{sec:discussion}

The main research question in Section~\ref{sec:intro} asks whether a multi-robot team can learn directed coordination preferences primarily from runtime co-occurrence statistics, without a task model, spatial priors, or offline training, while remaining adaptive under changing task distributions. The results presented in Section~\ref{sec:experiments} provide an affirmative answer across four fleet scales. MORPH learns who should coordinate with whom entirely from Jaccard co-assignment statistics, requires no positional input, operates from step zero without any offline phase, and degrades $3\times$ less than proximity-based methods under a spatial task distribution shift, and $2\times$ less than TSG. The learned preference structure is demonstrably task-semantic rather than spatial ($\rho = -0.111$, $p = 0.174$, n.s.), confirming that co-occurrence statistics alone are a sufficient signal for productive coordination structure to emerge.

The ablation and TSG comparisons together address RQ1. Co-occurrence statistics alone are not sufficient to infer productive coordination preferences and they must be accumulated, gated by delivery reward, stabilized through homeostasis, and structurally consolidated via link formation and pruning. At larger scales, the benefit of accumulated preference memory becomes clearer: MORPH remains above TSG at both Medium and Large scales while preserving a sparse coordination topology, whereas TSG keeps only a very small instantaneous co-assignment graph. This suggests that task similarity alone can be competitive in simpler settings, but persistent plasticity becomes more valuable as the team size grows and coordination dependencies become harder to recover from single-step co-assignment signals. The ablation results further support this interpretation: the full MORPH system leads all partial variants at every scale, removing the reward gate weakens the selectivity of R1, and removing neuromodulation produces the most consistent degradation. These results suggest that MORPH's performance comes from the interaction of reward-gated Hebbian strengthening, homeostatic degree regulation, structural link formation and pruning, and metaplastic control, rather than from a single heuristic.

Mechanistically, when demand shifts away from a previously active zone, the Jaccard signal on obsolete AGV--Picker pairs decreases, removing the synaptic input that had sustained those preferences through R1. The sliding threshold then helps stale, over-potentiated preferences decay instead of remaining locked into the previous demand regime. At the same time, the delivery-rate drop drives the metaplastic controller toward exploration: the formation threshold is reduced, new zone-relevant preferences can form through R3, and homeostatic regulation redistributes coordination degree as outdated links are pruned. The policy floor $W_{\text{floor}} = 0.25$ maintains a minimum response probability during this transition, so coordination does not collapse before the new structure is consolidated. These mechanisms address RQ2: the learned coordination structure is both persistent and revisable under non-stationary task distributions, with retention compounding across repeated regime cycles as the shift results show. MORPH therefore improves resilience not by replacing the topology after each shift, but by preserving reusable coordination preferences while allowing obsolete links to decay that is a pattern consistent with motor-learning models in which rapid adaptation arises from small, correlated connectivity changes rather than wholesale network replacement~\cite{feulner2022small}.

The most natural comparison is not with trained methods but with proximity as the incumbent training-free baseline. Proximity remains strong because spatial closeness is a useful inductive bias in TA-RWARE. It ties MORPH at Tiny and achieves higher raw throughput at Small, Medium, and Large. At Tiny scale MORPH ties Proximity exactly (62.4 deliveries each); at larger scales Proximity leads by 4--10 deliveries, but this gap must be interpreted carefully. Proximity therefore represents a strong upper-bound spatial heuristic rather than an uncalibrated field baseline. Under the task distribution shift experiment, which more closely models real deployment conditions, MORPH's advantage reverses: it degrades $3\times$ less than Proximity, recovers $4.7\times$ faster than Proximity and does so without prior positions calibration. The two methods have complementary strengths, and the choice between them should be driven by whether spatial priors are available and stable, not by raw throughput on a single fixed distribution.

The comparison with Full-Graph is equally important and less obvious. All-to-all coordination is consistently \emph{worse} than MORPH at every scale, and the gap grows with $N$ (from 4.6 deliveries at Tiny to 9.4 at Large). This is not a side effect of the soft-gating floor. It also reflects a genuine cost of coordination noise. When every AGV broadcasts to every Picker simultaneously, conflicting response signals grow as $O(N^2)$ and productive task-specific partnerships are obscured. MORPH's selectivity is not merely efficient, it is functionally beneficial, and the benefit compounds at scale. This finding has a direct implication for system design: in heterogeneous fleets, sparse learned coordination topology should be preferred over all-to-all broadcast even when bandwidth is not a constraint.

Taken together, these results position MORPH in a previously unoccupied region of the MRTA design space. Auction and optimisation methods require a utility function or cost matrix specified before deployment. Learning-based methods require millions of environment steps on a fixed distribution and cannot adapt after training. Proximity heuristics operate from step zero but encode a fixed spatial prior and require both position sensors and per-environment radius calibration. MORPH requires none of these. It starts from zero preferences, operates without spatial sensors, and adapts continuously when conditions change. The price of this flexibility is a cold-start period of approximately 50--100 steps during which preferences have not yet accumulated, which can be seen as a structural cost that is bounded, not a hyperparameter issue.

Several limitations bound the current contribution. First, MORPH is evaluated on a single benchmark environment; its behaviour on qualitatively different task structures, where, for example, spatial proximity and task co-assignment are not correlated, remains untested. Second, MORPH assumes that co-assignment statistics are a reliable proxy for productive partnerships. In environments where agents frequently co-assign without producing deliveries (e.g.\ due to congestion), the reward gate in R1 partially compensates but may not fully prevent noisy preference accumulation. A natural extension is to add eligibility traces to R1 so delayed deliveries can credit earlier co-assignments, following reward-modulated Hebbian models of delayed motor timing~\cite{kawai2023spatiotemporal}. Finally, all experiments are conducted in simulation; the gap between simulation and physical deployment, particularly regarding timing, communication latency, and sensor noise, has not been characterised.

 
\section{Conclusion}
\label{sec:conclusion}
 
We have presented MORPH, a training-free MRTA framework in which global allocation quality emerges from four interacting local plasticity rules applied to a directed pairwise preference matrix updated entirely from runtime co-occurrence statistics. MORPH requires no task model, no bid computation, no position sensors, and no offline training. Evaluated on the TA-RWARE robotic warehouse benchmark across four fleet scales ($N = 8$--$24$) and 5 seeds, MORPH achieves 110\% of all-to-all throughput at $N = 24$ while using only 21\% of possible coordination links, degrades $3\times$ less than proximity-based methods under spatial task distribution shift across six alternating demand phases, with recovery time decreasing across repeated regime cycles as retained preferences compound, and produces two allocation properties, cross-type preference dominance and progressive sparsification, that emerge without being programmed. These results position MORPH in a previously unoccupied region of the MRTA design space: a method that is simultaneously training-free, sensor-free, and adaptive, operating from the first episode in any environment without modification to the underlying robot controller.

The most direct extension is to evaluate MORPH on additional heterogeneous fleet benchmarks and physical robot platforms. On the algorithmic side, the plasticity parameters are currently set by hand;
meta-learning the plasticity rules themselves is a natural next step that would allow MORPH to adapt its own adaptation mechanism to the environment. A second direction is to extend the preference matrix
from pairwise to higher-order structures, capturing coordination relationships among groups of agents rather than pairs, which may be necessary for tasks requiring tight three-or-more-agent synchronisation. Finally, MORPH's self-organising coordination layer is in principle applicable to any existing robot controller; applying it to trained MARL policies combining the adaptability of online plasticity with the task performance of optimised policies is a promising direction toward systems that are both competent and resilient.

\bibliographystyle{splncs04}
\bibliography{ref}

\end{document}